\documentclass[runningheads]{llncs}

\usepackage[mobile]{eccv}

\usepackage{eccvabbrv}

\usepackage{graphicx}
\usepackage{booktabs}

\usepackage[accsupp]{axessibility}  % Improves PDF readability for those with disabilities.

\usepackage[pagebackref,breaklinks,colorlinks]{hyperref}
\usepackage{algorithm}
\usepackage{algpseudocode}
\usepackage{booktabs} % 

\begin{document}

\title{Powerful and Flexible: Personalized Text-to-Image Generation via Reinforcement Learning} 

\titlerunning{Personalized Text-to-Image Generation via Reinforcement Learning}

% Second Author\inst{2,3}\orcidlink{1111-2222-3333-4444} \and
% Third Author\inst{3}\orcidlink{2222--3333-4444-5555}}

\author{\small Fanyue Wei$^{1}$ \and
Wei Zeng$^{2}$ \and
Zhenyang Li$^{2}$ \and 
Dawei Yin$^{2}$ \and 
Lixin Duan$^{1}$ \and
Wen Li$^{1\dagger}$} %


\authorrunning{Wei et al.}

\institute{$^{1}$University of Electronic Science and Technology of China \\
\email{\{wfanyue, wzeng316, zhenyounglee, lxduan, liwenbnu\}@gmail.com} \\ 
$^{2}$Baidu Inc\\
\email{\{yindawei\}@acm.org}}

\maketitle

\begin{abstract}
  Personalized text-to-image models allow users to generate varied styles of images (specified with a sentence) for an object (specified with a set of reference images). While remarkable results have been achieved using diffusion-based generation models, the visual structure and details of the object are often unexpectedly changed during the diffusion process. One major reason is that these diffusion-based approaches typically adopt a simple reconstruction objective during training, which can hardly enforce appropriate structural consistency between the generated and the reference images. 
To this end, in this paper, we design a novel reinforcement learning framework by utilizing the deterministic policy gradient method for personalized text-to-image generation, with which various objectives, differential or even non-differential, can be easily incorporated to supervise the diffusion models to improve the quality of the generated images. Experimental results on personalized text-to-image generation benchmark datasets demonstrate that our proposed approach outperforms existing state-of-the-art methods by a large margin on visual fidelity while maintaining text-alignment. Our code is available at: \url{https://github.com/wfanyue/DPG-T2I-Personalization}.

  % \keywords{Personalized Text-to-Image Generation \and Reinforcement Learning \and Visual Fidelity}
\end{abstract}

\renewcommand{\thefootnote}{\fnsymbol{footnote}}
\footnotetext{$^{\dagger}$Corresponding author.}

\vspace{-0.6cm}
\section{Introduction}
\label{sec:intro}

% Background
Recent advances in text-to-image generation~\cite{ramesh2022hierarchical_dalle2, rombach2022high_stable_diffusion, saharia2022photorealistic_Imagen} exhibit the impressive ability to synthesize high-quality impressive images. Such models are robust and can generate images of diverse concepts in a wide variety of backgrounds and contexts. This opened a new area of research and innovation.

However, these generation models are uncontrolled and lack the ability to synthesize customized concepts from personal lives. For instance, it is not possible to query with a prompt/image from your own pets, friends, or personal objects and modify their poses, locations, styles, or backgrounds.

% Existing Methods
To achieve such customization, existing approaches~\cite{gal2022image_text_inversion, ruiz2023dreambooth, kumari2023multi_custom_diffusion} utilize a controlled fine-tuning mechanism which allows the possibility of embedding new concepts into the pre-trained text-to-image diffusion model. For example, Text-Inversion~\cite{gal2022image_text_inversion} personalizes image generation by learning a unique textural identifier of the new concept from a given set of images during fine-tuning. Then, the fine-tuned model is able to generate new variations of the input concept using a prompt containing the learned identifier. 
% Andrey \etal ~\cite{voynov2023p_p_plus} further improves the inversion method by injecting the learnable identifier embedding in each attention layer of the denoising U-Net. 
Besides, DreamBooth~\cite{ruiz2023dreambooth} fine-tunes the entire diffusion model to learn the personalized concept instead. It is additionally regularized by the super-class images to preserve the class-specific priors. 
In addition, Custom Diffusion~\cite{kumari2023multi_custom_diffusion} proposes to fine-tune the key and value parameters in each cross-attention layer to enhance computational efficiency. 
% ELITE~\cite{wei2023elite} proposes to learn an encoder for directly mapping visual concepts into textual embeddings. 
However, all these diffusion-based methods are trained by a simple reconstruction objective step-by-step which can hardly enforce appropriate visual consistency between the generated image and the reference images. 
%They generally follow the optimization process of diffusion and denoising with reconstruction loss at the each timestep thus weak in the constraints about the final generation results.

% Introduce our motivation
To this end, we design a novel framework for the task of text-to-image(T2I) personalization via reinforcement learning which is facilitated with various objectives, differential or even non-differential.  
There are existing text-to-image generation methods using reinforcement learning by utilizing human feedback~\cite{xu2023imagereward, fan2024reinforcement_dpok, christiano2017deep_learn_human_feedback, kirstain2024pick, clark2023directly_DRaFT}.
They usually use the policy gradient approach to incorporate aesthetic assessment or human preference as the reward for the general text-to-image generation to improve the image quality or text-alignment. While under the personalized setting, usually given only $4\sim6$ images depicting personalized concepts, it is hard to train an appropriate customized reward model. In our work, different from the existing reinforcement learning methods using human feedback reward for text-to-image generation, we explore several ways for text-to-image personalization to provide the suitable reward model for capturing the long-term visual consistency of the personalized subjects in diffusion model and rich supervision signals.

\begin{figure}[t]
  \centering
  \includegraphics[width=\linewidth]{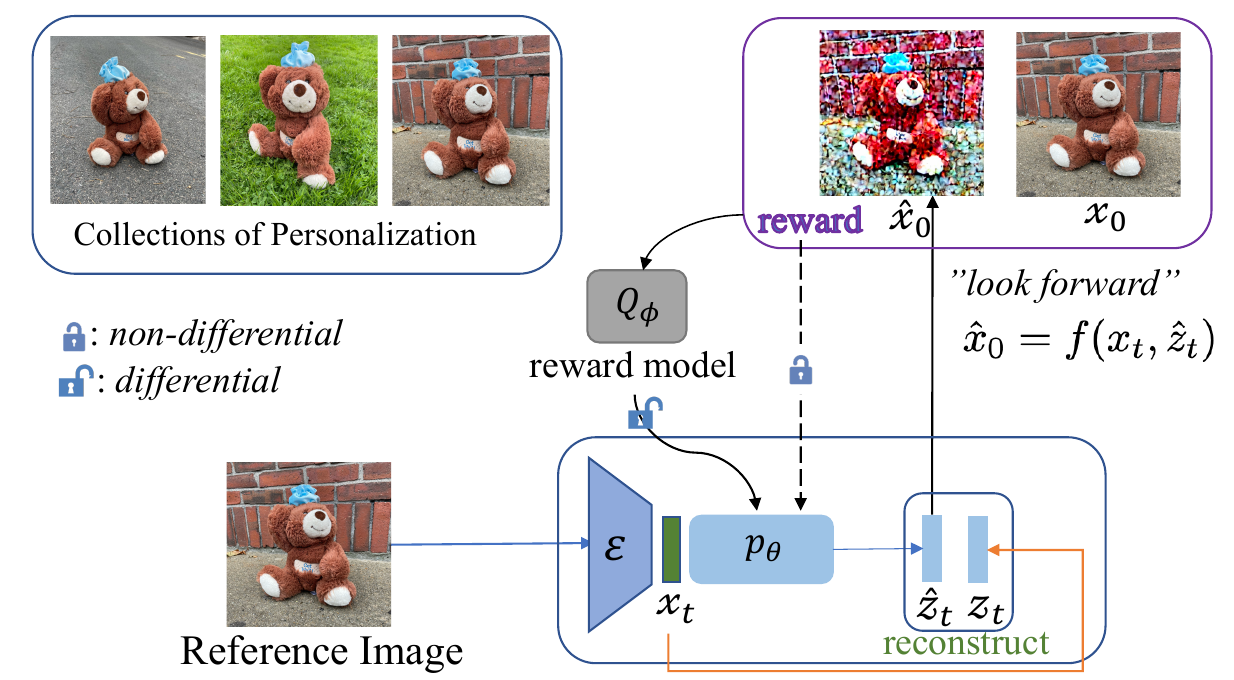}
  \caption{Our proposed framework utilizes the DPG algorithm to capture the visual consistency and supervises the generation model with flexible objectives, differential or even non-differential.} 
\label{fig_framework}
\end{figure}

%In this work, we propose a flexible framework that can facilitate different supervision for personalized text-to-image generation. As shown in Fig~\ref{figure_1_motivation}, in this framework, we employ the deterministic policy gradient (DPG) algorithm to fine-tune the diffusion model by introducing a differential specific reward function concerning personalized concepts, different from those human feedback reward in existing RL method for general text-to-image generation. 
In this study, we introduce a versatile framework designed to support various forms of supervision for personalized text-to-image generation. Illustrated in Fig.~\ref{fig_framework}, our framework utilizes the deterministic policy gradient (DPG) algorithm to fine-tune the diffusion models. This involves the incorporation of a specific differentiable reward function that considers personalized concepts. Based on this new framework, we further introduce two new losses to capture the long-term visual consistency for the text-to-image personalization task and enrich the supervision to improve the visual fidelity for personalization 

Experimental results show that our proposed approach surpasses existing state-of-the-art methods on multiple personalized text-to-image generation benchmarks by a large margin to preserve visual fidelity.  

In summary, Our main contributions are as follows:
\begin{itemize}
\item We design a novel framework for the task of text-to-image personalization via reinforcement learning. Especially, we regard the diffusion model as a deterministic policy and can be supervised by a learnable reward model for personalization.

\item With the flexibility of our proposed framework, We introduce two new losses to improve the quality of generated images to capture the long-term visual consistency for the personalized details and enrich the supervision for the diffusion model.
\item Experimental results on personalized text-to-image generation benchmarks demonstrate that our proposed approach surpasses existing state-of-the-art methods in visual fidelity.
% \item Besides, our framework can perform test-time optimization for the given prompt to improve the quality of generation.
\end{itemize}

\vspace{-0.5cm}
\section{Related Work}
\label{sec:related work}
\vspace{-0.2cm}

\subsection{Diffusion Models for Image Generation}
Diffusion-based image generation models~\cite{ramesh2021zero_dalle, ramesh2022hierarchical_dalle2, rombach2022high_stable_diffusion, saharia2022photorealistic_Imagen, dhariwal2021diffusion_classguidance, betker2023dalle3, nichol2021improved, ramesh2022hierarchical_dalle2, zhang2023forgedit, kawar2023imagic, podell2023sdxl} have developed rapid and impressive progress recently. DDPM~\cite{ho2020denoising_ddpm} first performs a noise diffusion during the forward process and denoises on a Markov process. Then, DDIM~\cite{song2020denoising_ddim} adopts an implicit estimation to accelerate the sample for image generation. 
As for text-to-image generation, there is also a huge progress. Imagen~\cite{saharia2022photorealistic_Imagen}, GLIDE~\cite{nichol2022glide}, Parti ~\cite{yu2022scaling}, Stable Diffusion~\cite{rombach2022high_stable_diffusion} and DALL·E ~\cite{betker2023dalle3} have all exhibited impressive results on image generation given a textual prompt. In particular, Stable Diffusion~\cite{rombach2022high_stable_diffusion} performs the diffusion process in the latent space, improving training and sampling efficiency significantly.

%-------------------------------------------------------------------------
\subsection{Personalized Text-to-Image Generation}
Personalized text-to-image generation~\cite{gal2022image_text_inversion, ruiz2023dreambooth, arar2023domain_agnostic, ruiz2023hyperdreambooth, gal2023encoder_e4t, hao2023vico, lee2024direct, arar2024palp} aims to adapt the pre-trained text-to-image generation model to learn a personalized concept from a given small set of images (\ie $4 \sim 6$ images) and modify its pose, style or context.

% inversion series
Text Inversion~\cite{gal2022image_text_inversion} personalizes the image generation by learning a unique textural identifier of the new concept in given images during fine-tuning. Then, the fine-tuned model is able to generate new variations of the input concept using a prompt containing the learned identifier. 
$\mathcal{P}+$~\cite{voynov2023p_p_plus} further improves the inversion method by injecting the learnable identifier into each attention layer of the denoising U-Net. In addition, NeTI~\cite{alaluf2023neural_neti} proposes to fuse the denoising process timestep on $\mathcal{P}+$~\cite{voynov2023p_p_plus} by introducing a neural mapper.
% 考虑去掉NeTI 

By contrast, DreamBooth~\cite{ruiz2023dreambooth} fine-tunes the entire diffusion model to learn the personalized concept. It is regularized by the super-class images to preserve the class-specific priors. Custom-Diffusion~\cite{kumari2023multi_custom_diffusion} proposes to only fine-tune the key and value parameters in the cross-attention layers to enhance computational efficiency. ELITE~\cite{wei2023elite} introduces to directly map the visual concepts into textual embeddings, by training a learnable encoder.

Besides, some works aim to provide a domain-specific text-to-image generator by utilizing a personalization encoder~\cite{arar2023domain_agnostic, chen2024subject_driven, gal2023encoder_e4t, jia2023taming, li2024blip, tewel2023key_lock, ma2023subject, shi2023instantbooth}. Given a single image and a prompt, these models enable to generate images within a specific class domain without fine-tuning on new input images.

% add content of pretrain on large domain to perform fast personalization??
Differently, this paper revisits the task of personalized text-to-image generation via reinforcement learning and reforms the learning paradigm into a deterministic policy gradient (DPG) framework.
 
\subsection{Reinforcement Learning for Text-to-image Generation}
\label{reinforcement_learning_reward}
Reinforcement learning method~\cite{rennie2017self, christiano2017deep_learn_human_feedback, kirstain2024pick, lee2023aligning, wu2023better, xu2023imagereward, lee2024parrot, liang2024rich, zhang2024large} has been explored for text-to-image generation using human preference as reward, they usually use aesthetic assessment or human preference for a given prompt.

DPOK~\cite{fan2024reinforcement_dpok} finetunes text-to-image generation model using
policy gradient with KL regularization using human feedback as reward, while DRaFT~\cite{clark2023directly_DRaFT} relies on the differentiable reward to propagate the reward function gradient across the sampling procedure in the denoising process.

They all finetune the diffusion model for the general text-to-image generation. While for the personalized setting, usually given the only $4\sim6$ images, it is hard to train an appropriate reward model. In our work, we explore several ways for the personalized text-to-image generation task with the specific reward model for the given personalized concepts based on DPG~\cite{silver2014deterministic, lillicrap2015continuous}.

\vspace{-0.3cm}
\section{Method}
\vspace{-0.3cm}

\subsection{Preliminaries}
\label{subsec: preliminaries}
Stable Diffusion (SD)~\cite{rombach2022high_stable_diffusion} is a latent text-to-image generation model based on DDPM~\cite{ho2020denoising_ddpm}. It contains a large autoencoder $\mathcal{E}$ which is pretrained to extract latents from images, and a corresponding decoder $\mathcal{D}$ to map the latents back to images for reconstruction $\mathcal{D}(\mathcal{E}(I)) \approx I$. SD performs the diffusion process on the latent space of the autoencoder $(\mathcal{E(\cdot), \mathcal{D}(\cdot)})$. Then, the text conditions $y$ can be injected into the diffusion process by cross-attention. 
Thus, the training objective of the diffusion model is:

\begin{equation}
    \mathcal{L}_{LDM}:= \mathbb{E}_{x \sim \sigma(x), y, \epsilon \sim \mathcal N} \left[ || \epsilon - \epsilon_{\theta}(x_{t}, t, \tau(y) ||^{2} \right],
\label{eq:ldm_loss}
\end{equation}
where $\mathcal{L}_{LDM}$ is a squared error loss, $\epsilon$ is the target noise, $\epsilon_{\theta}(\cdot)$ is a denoising network (\ie, U-Net) to predict noise adding to the latents, $t$ is timestep in diffusion process, $x_{t}$ is the noisy latents in timestep $t$, and $\tau(\cdot)$ is the pretrained CLIP~\cite{radford2021learning_clip} text encoder in Stable Diffusion~\cite{rombach2022high_stable_diffusion}. 
During inference, a random Gaussian noise $x_{T} \sim \mathcal{N}(0, 1)$ is iterative denoised to $x_{0}$, and the final image is obtained through the decoder $\hat{I}_{0} = \mathcal{D}(x_{0})$.

\vspace{-0.2cm}
% Reformulate T2I personalization of DPG framework
\subsection{DPG framework for T2I Personalization }
\label{sec: dpg for t2i personalization}

% \begin{figure}[t]
%   \centering
%   \includegraphics[width=\linewidth]{figures/Figure1.pdf}
%   \caption{The framework of our proposed DPG algorithm for text-to-image personalization} 
% \label{fig_framework}
% \end{figure}

Existing approaches~\cite{ruiz2023dreambooth, kumari2023multi_custom_diffusion, gal2022image_text_inversion} for text-to-image personalization follows the training procedure presented in Sec.~\ref{subsec: preliminaries}. The reconstruction loss is calculated during the diffusion process for $x_{t}$, thus cannot be directly used to optimize the final generation results $x_{0}$ by the visual details. Whereas RL technique can act as a flexible tool for optimization, and has achieved huge success in various fields due to its flexibility and powerful modeling capabilities. Inspired by this, we revisit the task of text-to-image personalization via reinforcement learning method. In particular, we treat the diffusion model as a deterministic policy and propose a flexible framework that can facilitate various supervision for personalized text-to-image generation with a learnable specific reward.   

Next, we outline the main formulation of our DPG framework. The deterministic policy applies the action that maximizes the Q-function $Q_{\phi}(\cdot)$, and the Q-function is assumed to be differentiable with respect to the action. 
In our framework, We regard the latent state, the timestep, and the encoded text condition $\{x_t, t, \tau(y)\}$ as the input, the predicted noise $z_t$ as the action, and the text-to-image generation model denoted by $\epsilon_{\theta}(x_t, t, \tau(y))$ as the policy. We define the policy function as follows,
\begin{equation}
\begin{aligned}
    \hat{z}_t &= \epsilon_{\theta}(x_{t},t,\tau(y)).
\label{eq:policy_function}
\end{aligned}
\end{equation}

At each timestep, the policy model $\epsilon_{\theta}(x_t, t, \tau(y))$ takes the latent $x_{t}$ of current timestep $t$ and the text condition $y$ as input and generates the action $\hat{z}_{t}$ during the training process.
 
As shown in Eq.~\ref{eq:expectation_reward}, the optimization purpose of the DPG framework is to maximize the expectation of the accumulated reward.
\begin{equation}
    \max_\theta { \mathbb{E}}\left[Q_\phi (x_{t},  \epsilon_{\theta}(x_{t},t,\tau(y)))\right],
\label{eq:expectation_reward}
\end{equation}
where $ Q_\phi (x_{t},  \epsilon_{\theta}(x_{t},t,\tau(y))) $ aims to calculate the cumulative reward when applies the action $z_t$ at the state $ \{x_t, \tau(y), t \}$.  

We optimize the Q-function based on the Carlo Monte Sampling Method to predict the accumulative reward. To achieve this objective, $Q_{\phi}(\cdot)$ is updated by the gradient descent algorithm using Eq.~\ref{eq:eq_gradient_q_phi} concurrently.
\begin{equation} 
    \min_{\phi} || Q_\phi (x_{t},  \epsilon_{\theta}(x_{t},t,\tau(y))) - \sum_{i=0}^{t} r(x_i, \tau(y), i) ||^{2},
\label{eq:eq_gradient_q_phi}
\end{equation}
where $r(x_i, \tau(y), i)$ denotes the reward on the timestep $i$.

While in the diffusion and denoising process, the policy model $\epsilon_{\theta}(\cdot)$ is optimized to minimize Eq.~\ref{eq:ldm_loss} in one timestep. Therefore, minimizing the reconstruction loss encourages policy model $\epsilon_{\theta}(\cdot)$ to make rewards (\ie, minimizing Eq.~\ref{eq:ldm_loss} equals to maximizing Eq.~\ref{eq:expectation_reward} for diffusion model).
% Here, the reconstruction loss can work as a kind of reward function, since the reward is to encourage policy model $p_{\theta}$ take the maximum value to make action (\ie,
 In this case, the reward function is obtained as follows,
\begin{equation}
    r(x_{t}, t, \tau(y))=  || \epsilon - \epsilon_{\theta}(x_{t}, t, \tau(y)) ||^{2}.
\label{eq:l2_noise_reward}
\end{equation}

$Q_{\phi}(\cdot)$ works to predict the one step immediate reward in Eq.~\ref{eq:l2_noise_reward} for the diffusion model $\epsilon_{\theta}(\cdot)$ given $x_{t}$ to estimate the noise in timestep $t - 1$. 

Therefore, as shown in Fig.~\ref{fig_framework}, the entire DPG framework for text-to-image personalization can be optimized by Eq.~\ref{eq:eq_gradient_q_phi} to train the diffusion model $\epsilon_{\theta}(\cdot)$ concurrently with $Q_{\phi}(\cdot)$. The algorithm pseudo code is presented in Algorithm~\ref{alg:dpg_algorithm}.

\begin{algorithm}
\caption{DPG Framework for T2I Personalization}
\label{alg:dpg_algorithm}
    \begin{algorithmic}[1]
        \State Input:  Policy function $\epsilon_\theta(\cdot)$, Q-function $Q_\phi(\cdot)$, the timestep $T$ and the text condition $y$
        \Repeat
        \State Randomly choose an image from the set of reference images
        \State Randomly sample $t \sim \{ 0, \cdots, T-1 \}$
        \State Process $t$ step diffusion process, obtain the latent state $x_t$
        \State Calculate the reward $r$ based on the definition 
        \State Obtain the accumulative reward as the target  \[ \sum_{i=0}^{t} r(x_i, \tau(y), i)\]
        \State Update the Q-function  parameters $\phi$ by one step of gradient descent using 
        \[
            \nabla_{\phi} || Q_\phi (x_{t},  \epsilon_{\theta}(x_{t},t,\tau(y))) - \sum_{i=0}^{t} r(x_i, \tau(y), i) ||^{2} 
        \]
        \State Update the diffusion model $\theta$ by  one step of gradient ascent using 
        \[
            \nabla_{\theta} Q_\phi (x_{t},  \epsilon_{\theta}(x_{t},t,\tau(y)))
        \]
        \Until{End Training}
    \end{algorithmic}
\end{algorithm}

% \begin{algorithm}
% \caption{DPG Framework for T2I Penalization}
% \begin{algorithmic}[2]
% \State Input:  policy function $p_\theta$ and Q-function parameters $Q_\phi$
% \Repeat
% \Procedure{DPG}{$a,b$}\Comment{calculate }
%     \State $r \gets a \bmod b$
%      \State $r \gets a \bmod b$
%      \State Compute targets
%       \State \ \ \ \ \ \ $y = || \epsilon_{\theta}((z_{t}, t, \tau_{\theta}(y)) - \epsilon ||^{2}$
%     \While{$r \neq 0$}\Comment{do the loop 0}
%         \State $a \gets b$
%         \State $\nabla_{\phi} \frac{1}{B} \sum_{B} || Q_{\phi} - y ||^{2}$
%         \State $r \gets a \bmod b$
%          \State $\nabla_{\theta} \frac{1}{B} \sum_{B}  Q_{\phi}$
%         \State $b \gets r$
%         \State $r \gets a \bmod b$
%     \EndWhile
%     \State \textbf{return} $b$\Comment{res $b$}
% \EndProcedure
% \end{algorithmic}
% \end{algorithm}

\vspace{-1cm}
\subsection{Learning to "Look Forward"}

\begin{figure}[ht]
  \centering
  \includegraphics[width=\linewidth]{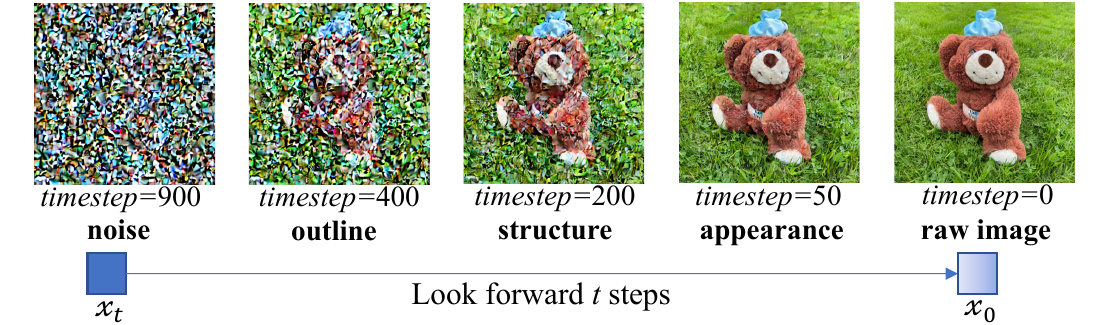}
  \caption{During the denoising process, in the early timesteps ($t \approx T$), the diffusion model attempts to represent the outline and structure of the subject, whereas in the later steps ($t \approx 0$), the model focuses on the visual details.} 
\label{fig:figure_timestep}
\end{figure}

Existing methods~\cite{ruiz2023dreambooth, gal2022image_text_inversion, kumari2023multi_custom_diffusion} follow the paradigm that the diffusion model is optimized by the reconstruction loss during the diffusion process step-by-step, thus cannot be directly optimized with the final generation images by the visual details. However, as shown in Fig.~\ref{fig:figure_timestep} intuitively, the denoising process is guided by different nature implicitly. At the different timesteps, the generation model focuses on the different features, in the early timesteps ($t \approx T$), the diffusion model attempts to recognize the outline of the subject and determine the structure of the subject, while in the later steps ($t \approx 0$), the model focuses on the visual fine details. Thus, we argue for taking advantage of the "look forward" of reinforcement learning to implicitly guide the generation model to capture the long-term visual consistency. This approach encourages the generation model to focus on the different features at different timesteps, improving the visual consistency of the personalized subject.

\vspace{-0.5cm}

Since the reward in Eq.~\ref{eq:l2_noise_reward} represents the one-step direct reward, we aim to leverage the nature of the diffusion process to "look forward" to $\hat{x}_{0, t}$. In the diffusion process, the Gaussian noise $z_{t}$ is added into the initial latent $x_{0}$ at timestep $t$ as follows,
\begin{equation}
    x_{t} = \sqrt{\bar{\alpha}_{t} } x_{0} + \sqrt{1 - \bar{\alpha}_{t} } z_{t}.
\label{eq:x_t_from_x0}
\end{equation}

During the denoising process, the policy diffusion model $\epsilon_{\theta}(z_{t}, t, \tau(y))$ aims to estimate the noise $\hat{z}_{t}$ from $x_{t}$. Consequently, we can obtain the $\hat{x}_{0,t}$ in Eq.~\ref{eq:x_hat_0}, derived from Eq.~\ref{eq:x_t_from_x0} at the given timestep $t$. 
\begin{equation}
    \hat{x}_{0, t} =  \frac{1}{\sqrt{\bar{\alpha}_{t} }} (x_{t} - \sqrt{1 - \bar{\alpha}_{t} } \hat{z}_{t}).
\label{eq:x_hat_0}
\end{equation}

After obtaining the final generation results $\hat{x}_{0, t}$, our purpose can transition from the step-by-step reconstruction of the diffusion noise to direct comparison of final generation results between $x_{0}$ and $\hat{x}_{0, t}$ at timestep $t$.

To achieve the objective of "looking forward" to implicitly guide the focus at different denoising states, the reward function in Eq.~\ref{eq:l2_noise_reward} can be rewritten between $x_{0}$ and $\hat{x}_{0, t}$ as follows,
\begin{equation}
\begin{split}
    \mathcal{L} &= || \hat{x}_{0, t} - x_{0} ||^{2} \\
                &= || \frac{1}{\sqrt{\bar{\alpha}_{t} }} (x_{t} - \sqrt{1 - \bar{\alpha}_{t} } \hat{z}_{t})  - \frac{1}{\sqrt{\bar{\alpha}_{t} }} (x_{t} - \sqrt{1 - \bar{\alpha}_{t} } z_{t})||^{2} \\
                &= \frac{1 - \bar{\alpha}_{t}}{\bar{\alpha}_{t}} || \hat{z}_{t}  - z_{t} ||^{2}.
\end{split}
\label{eq:x_0_loss}
\end{equation}

With the optimization objective on $\hat{x}_{0, t}$ in Eq.~\ref{eq:x_0_loss}, our DPG framework can learn to look forward from $x_t$ to $\hat{x}_{0,t}$ to acquire the implicit guidance at different timestep $t$ to enforce appropriate long-term structural consistency between the generated image and the reference images. Thus, the reward function in Eq.~\ref{eq:x_0_loss} can be used to update $Q_{\phi}(\cdot)$ at step $8$ in Algorithm~\ref{alg:dpg_algorithm}.

In addition, to make utilization of the prior that the Gaussian noise is added gradually over $t$ steps and denoised step-by-step, we accumulated the reward from $t$ to $0$ in Eq.~\ref{eq:q_phi_gamma} to make consistency with the denoising process rather than calculating one-step direct rewards,
\begin{equation} 
     Q_\phi (x_{t},  \epsilon_{\theta}(x_{t},t,\tau(y))) = \mathcal{L}(x_t, \tau(y), t) +  \gamma Q_\phi (x_{t-1},  \epsilon_{\theta}(x_{t-1},t-1,\tau(y))).
\label{eq:q_phi_gamma}
\end{equation}

% \begin{equation} 
%      Q_\phi (x_{t},  \epsilon_{\theta}(x_{t},t,\tau(y))) = r(x_i, \tau(y), i) +  \gamma Q_\phi (x_{t-1},  \epsilon_{\theta}(x_{t-1},t-1,\tau(y)))
% \end{equation}

% The Q-function $Q_{\phi}(\cdot)$ in Eq.~\ref{eq:q_phi_gamma} denotes the accumulated expectation of the history reward.
% Here, we rewrite the reward function with the timestep discount rate to consider the effect of "Look forward" to $\hat{x}_{0, t}$ in the diffusion process. We denote the discount rate with $\gamma$, then we modify the Eq.~\ref{eq:expectation_reward} as follows, where $B$ denotes the batch size.

% \begin{equation}
%    \nabla_{\phi} \frac{1}{B} \sum_{B} Q_\phi (x_{t}, \epsilon_{\theta}(x_{t},t,\tau(y)))
% \label{eq:expectation_reward_t}
% \end{equation}

The pseudo code is included in the supplementary materials.

\vspace{-0.2cm}
\subsection{Learning Complex Reward}
\label{sec: Learning Complex Reward}

\vspace{-0.7cm}
\begin{figure}[ht]
  \centering
  \includegraphics[width=\linewidth]{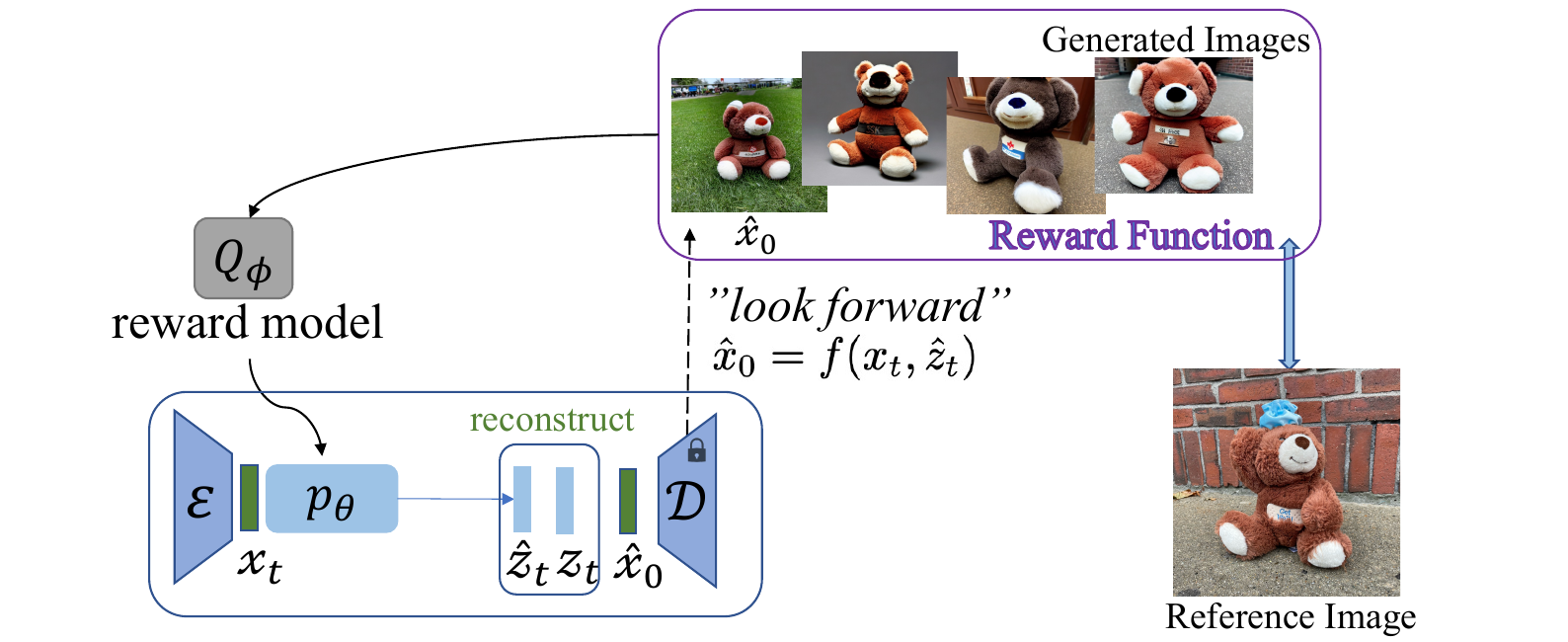}
  \caption{Our proposed framework of DPG equipped with "looking forward"  can further introduce more flexible supervision with a learnable reward model for the personalized generation model (\eg, Stable Diffusion).} 
\label{fig:complex_objective_reward}
\end{figure}
\vspace{-0.4cm}

With our proposed DPG framework equipped with "Look Forward" for text-to-image personalization, we are capable to incorporate various complex objectives, differential or even non-differential, to learn a specific reward model $Q_{\phi}(\cdot)$ to supervise the generation models to improve the quality of generated images. As shown in Fig.~\ref{fig:complex_objective_reward}, the aforementioned strengths arise from two aspects: one is that "Look forward" allows us to optimize the model based on the final generation results, other is that the learnable $Q_{\phi}(\cdot)$ for personalized subjects facilities complex supervision.

As the self-supervised learning method DINO~\cite{caron2021emerging_dino} encourages the unique visual features for personalization~\cite{ruiz2023dreambooth}, in this work, we select the DINO similarity as the representative reward function. 

With our specific reward model $Q_{\phi}(\cdot)$ for the given collection of personalized reference images $I(\cdots)$, we can simply incorporate the DINO reward for the diffusion model in our DPG framework.

To adopt the DINO similarity as reward, we utilize $\mathcal{D}(\cdot)$ to decode the final generation results $\hat{x}_{0, t}$ to the final generated image $\hat{I}$ by using $\hat{I}=\mathcal{D}(\hat{x_{0, t}})$. Then, we obtain the image embeddings from DINO image encoder $\mathcal{X}(\cdot)$ by $\kappa = \mathcal{X}(I)$. Thus, the reward function $r(\cdot)$ to be estimated by $Q_{\phi}(\cdot)$ can be formulated as follows,
\begin{equation}
    r(x_{t}) = - (1 - \hat{\kappa} \cdot \kappa),
\label{dino_reward_function}
\end{equation}
where $\hat{\kappa}$ is embeddings of the generated image extracted from $\mathcal{X}(\cdot)$, while $\kappa$ refers to the embeddings of the reference image.

% combine dino监督 + reconstruction loss  % 没有强调 differential

Then, we can inject the supervision of unique visual features of the personalized subjects with the original reconstruction reward, as in Eq.~\ref{eq:l2_noise_reward}. Thus we adopt the combination of both the DINO reward and reconstruction reward in Eq.~\ref{eq:l2_noise_reward} to work as the complex reward function, controlled with a weight $\lambda$, where $B$ denotes batch size. The gradient at step $9$ in Algorithm~\ref{alg:dpg_algorithm} can be refined as follows,
\begin{equation}
   \nabla_{\theta} \frac{1}{B} \sum_{B}  (\lambda Q_\phi (x_{t},  \epsilon_{\theta}(x_{t},t,\tau(y))) + ( -|| \epsilon - \epsilon_{\theta}(x_{t}, t, \tau(y))||^{2})).
\end{equation}

With our flexible DPG framework for reward, the reward can be differential to any other metric that can reflect on the generated image along with the reference personalized images.

% \subsection{Test-time Optimization for Any prompt}

% Utilizing the pretrained $Q_{\phi}(\cdot)$ in Sec.~\ref{sec: Learning Complex Reward}, as we find that, the $Q_{\phi}$ which adopt DINO similarity between the raw image and the generated image can learn some implicit unique features such as structure, color for personalization subjects. Therefore, with the differential reward function and the advantage of "look forward" of our DPG framework, we can still optimize the generation model during inference time.

% Given a textual prompt $y$ and a random initial noise $x_{T}$, $x_{t}$ is obtained during the denoising process in diffusion model. Then, we obtain $\hat{x}_{0, t}$ corresponding to the given prompt $y$, which means the latents in timestep $t$. By the pretrained $Q_{\phi}$ proposed in Sec~\ref{sec: Learning Complex Reward}, we can obtain the reward at timestep $t$, which represents the DINO distance with respect to the generated image and the real images.

% Therefore, we utilize the $Q_{\phi}$ as the supervision for the inference time, and work as the differential reward function to guidance the policy diffusion model $p_{\theta}$. By simply maximize

% \begin{equation}
%     \max_\theta { \mathbb{E}}\left[Q_{\phi, t} (\hat{x}_{0, t},\tau(y) ,t, z_{t})\right].
% \label{expectation_reward}
% \end{equation}

% Therefore, given a complex new text condition, with our framework, we can make it differential to optimize the policy diffusion model $p_{\theta}$ to improve the quality of generation.

\vspace{-0.5cm}
\section{Experiments}
\vspace{-0.3cm}

\subsection{Experimental Setup}

\noindent\textbf{Datasets.} We adopt the DreamBooth benchmark~\cite{ruiz2023dreambooth} to evaluate our DPG framework for text-to-image personalization. The dataset comprises $30$ concepts across $15$ different categories. Among these, $9$ subjects belong to live pets (\ie, dogs and cats). The remaining $21$ subjects pertain to various objects such as backpacks, cars and \etc. The dataset contains $4 \sim 6$ of images per concept, each captured under differing conditions, in various environments, and from multiple perspectives. Moreover, the dataset also contains $25$ challenging prompts to evaluate the text-to-image personalization methods. Besides, we also conduct experiments on Custom benchmark~\cite{kumari2023multi_custom_diffusion}.

\begin{table}[ht]
\caption{Quantitative comparisons with existing methods on DreamBooth benchmark.}
\centering
\begin{tabular}{cccc}
  
  \specialrule{.12em}{.1em}{.15em} 
  Method & DINO  & CLIP-I & CLIP-T  \\
  \specialrule{.1em}{.1em}{.15em} 
  Custom Diffusion~\cite{kumari2023multi_custom_diffusion} & 0.649 & 0.712 & 0.321 \\
  \specialrule{.06em}{.06em}{.15em} 
  Custom Diffusion w/ Our DINO reward & 0.640 & 0.715 & 0.320 \\
  Custom Diffusion w/ Our Look Forward & 0.669 & 0.728 & \textbf{0.322} \\
  \specialrule{.1em}{.15em}{.15em} 
  DreamBooth~\cite{ruiz2023dreambooth} & 0.694 & 0.762 & 0.282 \\
  \hline
  DreamBooth w/ Our DINO reward & 0.723 & 0.783 & 0.270 \\
  DreamBooth w/ Our Look Forward & \textbf{0.738} & \textbf{0.797} & 0.269 \\
  \specialrule{.12em}{.15em}{.15em} 
\end{tabular}
\label{tab:quantitative_comparisons}
\end{table}

\noindent \textbf{Evaluation:} Following existing text-to-image personalization methods~\cite{ruiz2023dreambooth, gal2022image_text_inversion, kumari2023multi_custom_diffusion}, we evaluate our proposed approach using \emph{Image-Alignment} and the \emph{Text-Alignment}. The \emph{Image-Alignment} measures the subject fidelity in the generated images while the \emph{Text-Alignment} aims to evaluate the similarity between the generated images and the given prompt.

For \emph{Image-Alignment}, we adopt DINO~\cite{caron2021emerging_dino} and CLIP-I~\cite{radford2021learning_clip}. The objective of self-supervised training in DINO~\cite{caron2021emerging_dino} is to encourage the discrimination of unique features of the subject, while CLIP-I~\cite{radford2021learning_clip} may focus on the semantic feature space such as color. Both DINO~\cite{caron2021emerging_dino} and CLIP-I~\cite{radford2021learning_clip} calculate the average pairwise cosine similarity between the extracted embeddings by the image encoder of the generated image and the reference image. For the \emph{Text-Alignment}, we calculate the similarity of the CLIP embeddings between the textual prompt and the generated image. Following existing methods~\cite{ruiz2023dreambooth, caron2021emerging_dino, kumari2023multi_custom_diffusion}, we adopt ViT-B/32
for the CLIP model and ViT-S/16 for the DINO model to extract visual and textual features.

\noindent \textbf{Implementation Details}. We adopt DreamBooth~\cite{ruiz2023dreambooth} as our baseline method. Since the official code of DreamBooth~\cite{ruiz2023dreambooth} is not publicly available, we use the implementation in the popular "diffusers" library\footnotemark[1] and the pretrained Stable Diffusion V1.4 for all compared methods for fair comparison. 
\footnotetext[1]{https://github.com/huggingface/diffusers}
For Custom Diffusion~\cite{kumari2023multi_custom_diffusion}, we reproduce their methods using their official code and the default hyperparameters on the DreamBooth benchmark. The resolution of generated images is $512\times512$. All experiments are performed on a 32G V100 card. More implementation details are included in the supplementary materials.

\subsection{Qualitative Results.} 
\label{sec: Qualitative Results}

\begin{figure}[ht]
  \centering
  \includegraphics[width=\linewidth]{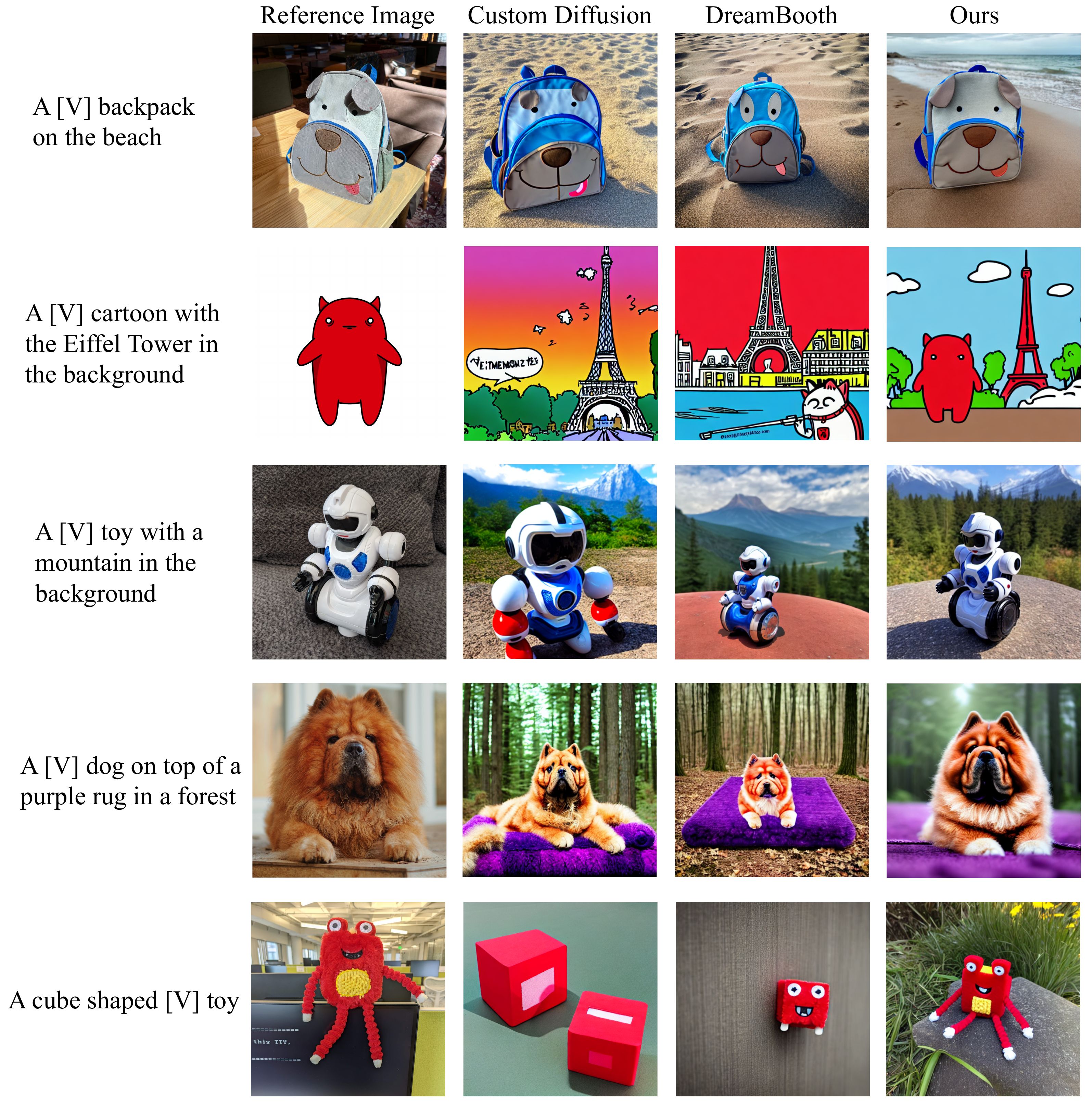}
  \caption{In this figure, we present the reference images alongside the images generated by Custom Diffusion, DreamBooth and our method. As demonstrated, given the challenging textual prompts, the images generated by Ours best preserve the high fidelity of the personalized attributes, including color, expressions, texture and \etc.} 
\label{fig:figure_showcase}
\end{figure}

We present the visualization results for qualitative comparisons. As shown in Fig.~\ref{fig:figure_showcase}, our method is capable to capture the visual details (\ie, color, texture, poses and \etc ) of the personalized subjects with various prompts while following the textual prompts faithfully. For example, the generated images by \emph{Ours} best capture the color of the reference images for the "backpack" and the shape of "cartoon" than the compared methods. The qualitative results demonstrate that our methods achieve better fidelity while preserving text-alignment. Additionally, our flexible DPG framework allows our method to be easily extended to other rewards for corresponding purposes. More visualization examples are included in the supplementary materials.

\subsection{Quantitative Results} 
  To validate the effectiveness of our proposed DPG framework, we compare our approaches (including the "Look Forward" reconstruction reward and the DINO reward) with several state-of-the-art methods. The compared methods include Custom Diffusion~\cite{kumari2023multi_custom_diffusion} and DreamBooth~\cite{ruiz2023dreambooth}. As illustrated in Table~\ref{tab:quantitative_comparisons} and~\ref{tab:custom_dataset}, our methods achieve the highest image-alignment on both DINO and CLIP-I evaluation metrics while preserving the text-alignment on CLIP-T metric.

 Especially, DreamBooth equipped with our proposed DPG framework improves the visual fidelity by a large margin $2.1\%$\ on DINO and $1.8\%$ on CLIP-I than the DreamBooth baseline methods. For text-alignment, there is an intrinsic trade-off between the text-alignment and image-alignment~\cite{kumari2023multi_custom_diffusion}. The current challenge of text-to-image personalization lies in improving visual consistency with the reference images, while the maintenance of text-alignment is primarily handled by the base T2I generation model such as Stable Diffusion and the text encoder. Our proposed methods outperform the compared methods on visual fidelity and preserve the text-alignment.
% In addition, the text-alignment is calculated with the CLIP encoder, which may also can not accurately measure the text-alignment.
For the Custom Diffusion~\cite{kumari2023multi_custom_diffusion} baseline, this approach only fine-tunes the parameters of cross-attention layers (between the text and image) while not emphasizing the visual embeddings, thus achieving better text-alignment performance yet poor image-alignment than the compared methods of the DreamBooth baseline.

\noindent \textbf{User Study.} We conduct a user study to compare our proposed approach with DreamBooth~\cite{ruiz2023dreambooth} to evaluate the human preference with the generation results for both the image-alignment and text-alignment. We provide three randomly selected personalized datasets for the user study. The participants are required to compare the generated images, which are the best results selected from eight images generated by different methods with the same random seed, given both the prompt and reference images. For the image fidelity, the participants are required to choose which method best preserves personalized visual consistency with reference images and which is most consistent with the prompt for text-alignment. As illustrated by Table~\ref{tab:quantitative_comparisons_user_study}, our method preserves image fidelity better than the compared method by a large margin while achieving comparable performance of text fidelity, which indicates that our approach can generate images that are more appealing to human preference. \\
\noindent \textbf{Computation Cost.} Our lightweight reward model operates on the latent space, thus introducing negligible computational cost. The number of trainable parameters is $0.26M$ \emph{versus} $859.40M$ trainable U-Net parameters.

Besides, we provide more analysis in the supplementary materials.

\begin{table}[ht]
\centering
\begin{minipage}[t]{0.45\textwidth}
\centering
\caption{Evaluation on Custom Benchmark.}
\footnotesize
\centering
\setlength{\tabcolsep}{0.3mm}{
    \begin{tabular}{cccc}
      \hline
      Method & DINO & CLIP-I& CLIP-T \\
      \hline
      DreamBooth & 0.640 & 0.737  &  0.309\\
      DreamBooth w/ LF & \textbf{0.680} & \textbf{0.773} & 0.303 \\
      DreamBooth w/ DINO & 0.653 & 0.753 &  \textbf{0.310}\\
      \hline
    \end{tabular}
}
\label{tab:custom_dataset}
\end{minipage}
\hfill
\begin{minipage}[t]{0.45\textwidth}
\centering
\footnotesize
    \caption{Quantitative comparisons of User Preference}
\setlength{\tabcolsep}{0.6mm}{
    \begin{tabular}{cccc}
      \hline
      Method & Ours & DreamBooth & Similar \\
      \hline
      Image Fidelity &  55.1\% & 12.0\% & 32.9\%\\
      Text Fidelity & 19.6\% & 20.4\% & 60.0\% \\
      \hline
    \end{tabular}
}
\label{tab:quantitative_comparisons_user_study}
\end{minipage}
\end{table}

\subsection{Ablation Studies}
\begin{figure}[ht]
  \centering
  \includegraphics[width=\linewidth]{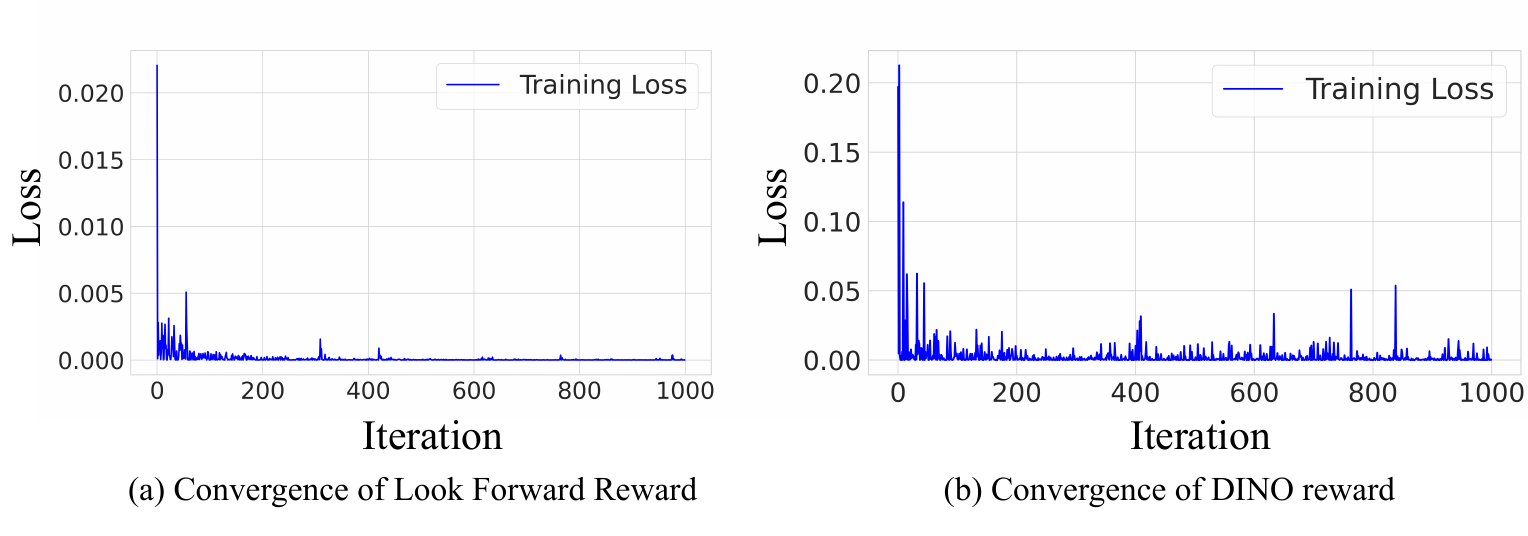}
  \caption{The convergence of the Q-function is illustrated in the subfigures. Subfigure (a) presents the training loss of the Q-function for the reconstruction reward, while subfigure (b) relates to the DINO reward.} 
\label{fig:loss_Q}
\end{figure}

We conducted ablation studies to verify the different components of our DPG framework, including the discount rate and the weight of the dino-reward. In addition, Fig.~\ref{fig:loss_Q} illustrates that our proposed reward model is easy to converge for both "Look Forward" reward and DINO reward.

\noindent \textbf{Effectiveness of the discount rate: } We adapt the discount rate of reinforcement learning to verify the sensitiveness of our framework. We conduct with different gamma on the "clock" dataset to demonstrate the robustness of our approach. As shown in Table~\ref{tab:ablation_discount_rate}, even with different discount rates, our methods still maintain high fidelity.

\vspace{-0.5cm}

\begin{table}[ht]
\centering
\begin{minipage}[t]{0.45\textwidth}
\footnotesize
\caption{Sensitivity of discount rate $\gamma$}
\vspace{-0.3cm}
\setlength{\tabcolsep}{0.8mm}{
    \begin{tabular}{cccc}
      \hline
     $\gamma$ & DINO & CLIP-I & CLIP-T \\
      \hline
      DB~\cite{ruiz2023dreambooth} & 0.644 & 0.707 & 0.239 \\
      w/o $\gamma$ & 0.727 & 0.761 & 0.209 \\
      $\gamma=0.9986$ & 0.704 & 0.743 & 0.213 \\
      \hline
    \end{tabular}
}
\label{tab:ablation_discount_rate}
\end{minipage}
\begin{minipage}[t]{0.45\textwidth}
\caption{Sensitivity of weight $\lambda$}
\vspace{-0.3cm}
\centering
\footnotesize
\setlength{\tabcolsep}{0.8mm}{
    \begin{tabular}{cccc}
      \hline
       $\lambda$ & DINO & CLIP-I & CLIP-T \\
      \hline
      DB~\cite{ruiz2023dreambooth} & 0.644 & 0.707 & 0.239 \\
      $\lambda = 0.1$ & 0.704 & 0.743 & 0.213 \\
      $\lambda = 1$ & 0.727 & 0.746 & 0.211 \\
      \hline
    \end{tabular}
    }
\label{tab:ablation_loss_weight}
\end{minipage}
\end{table}

\begin{figure}[ht]
  \centering
  \includegraphics[width=0.8\linewidth]{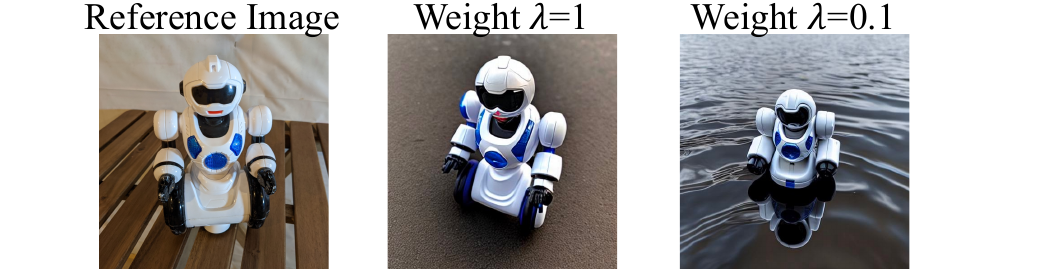}
  \caption{As illustrated in the figure, with the textual prompt "A [V] toy floating on the water" to generate images, increasing the weight of DINO reward may preserve the attributes of the personalized subject but damage the ability of text-alignment.} 
\label{fig:loss_weight_ablation}
\end{figure}

\noindent \textbf{Different weight for DINO reward $\lambda$: } We evaluate the sensitivity of our DPG framework by ablating the weight $\lambda$ for the DINO reward. We conduct experiments using different $\lambda$ values of $0.1$ and $1$ on the "robot\_toy" dataset to demonstrate the robustness of our approach. As shown in Table~\ref{tab:ablation_loss_weight}, increasing the weight $\lambda$ for the DINO reward from $0.1$ to $1$ improves the visual fidelity marginally. The DINO metric steeply increases from $0.644$ to $0.727$, and CLIP-I from $0.707$ to $0.746$. However, a trade-off may exist between the high visual reconstruction of personalization and text-alignment. As shown in Table~\ref{tab:ablation_loss_weight}, CLIP-T drops from $0.239$ to $0.211$. This indicates that as the DINO reward increases, the generation model tends to emphasize visual fidelity, which may potentially compromise the text-alignment ability. We present the images generated by the generation model with two different $\lambda$ weights in Fig.~\ref{fig:loss_weight_ablation}. 

As shown in the figure, the robot generated by the generation model with higher $\lambda$ overemphasizes better visual consistency with the reference images but follows the textual prompt less faithfully.

\vspace{-0.3cm}
\section{Conclusion}
\vspace{-0.2cm}
We design a novel framework for text-to-image personalization via reinforcement learning. Especially, we treat the diffusion model as a deterministic policy that can be supervised by a learnable reward model for personalization. With the flexibility of our framework, we introduce two new losses to improve the quality of generated images. The proposed method is capable to capture the long-term visual consistency of personalized details and enrich the supervision of the diffusion model. Experiments on several benchmarks demonstrate that our approach surpasses existing methods in visual fidelity while preserving text-alignment.

\noindent \textbf{Limitations: }In some cases, our framework equipped with such baselines (\eg, DreamBooth) may overemphasize the visual fidelity. The issue can be alleviated with a stronger text encoder or by resorting to baselines which balance the alignment between image and text. Moreover, we will further design the text-alignment related reward with our DPG framework to improve text-alignment.

\noindent \textbf{Social Impact: }Our methods can synthesize some fake images with personalized subjects such as human face or private pets, which may increase the risk of privacy leakage and portrait forgery. Therefore, users intending to use our technique should apply for authorization to use the respective personalized images. Nevertheless, our approach can serve as a tool for AIGC to create imaginative images for entertainment purposes.

\bibliographystyle{splncs04}
\bibliography{egbib}
\end{document}